# Adaptive Adversarial Training Does Not Increase Recourse Costs

Ian Hardy
ihardy@ucsc.edu
University of California, Santa Cruz
Santa Cruz, California, USA

Jayanth Yetukuri
jayanth.yetukuri@ucsc.edu
University of California, Santa Cruz
Santa Cruz, California, USA

Yang Liu*
yangliu@ucsc.edu
University of California, Santa Cruz
Santa Cruz, California, USA

## ABSTRACT

Recent work has connected adversarial attack methods and algorithmic recourse methods: both seek minimal changes to an input instance which alter a model's classification decision. It has been shown that traditional adversarial training, which seeks to minimize a classifier's susceptibility to malicious perturbations, increases the cost of generated recourse; with larger adversarial training radii correlating with higher recourse costs. From the perspective of algorithmic recourse, however, the appropriate adversarial training radius has always been unknown. Another recent line of work has motivated adversarial training with adaptive training radii to address the issue of instance-wise variable adversarial vulnerability, showing success in domains with unknown attack radii. This work studies the effects of adaptive adversarial training on algorithmic recourse costs. We establish that the improvements in model robustness induced by adaptive adversarial training show little effect on algorithmic recourse costs, providing a potential avenue for affordable robustness in domains where recoursability is critical.

## 1 INTRODUCTION

The adoption of Machine Learning (ML) in consequential environments motivates the provision of instructions to adversely-affected users on actions they can take to alter a model's decision. For example, in the lending domain, if a classifier decides to deny an applicant, there should be a mechanism for providing a feasible set of actions the applicant can take to be approved. This instructive information is desirable as opaque self-learning systems inform more and more of our society's decision-making, for both trust and accountability. The ability to obtain a desired outcome from a known model, the actionable set of changes that users can make to improve their qualification, or the systematic process of reversing unfavorable decisions is defined as "algorithmic recourse," or simply "recourse" [12]. These what-if scenarios are also often referred to as "counterfactual explanations." Importantly, the explicitly stated goal of recourse is to find actions with minimal cost to the user [24].

Simultaneously, it has been observed that many neural networks can be easily "fooled" by introducing small changes to input features that may seem imperceptible. [22] first proposed the concept of "adversarial examples": by adding small perturbations to an input sample, models obtain incorrect classification results with high confidence scores. These are sometimes referred to as "evasion attacks" [5]. [22] also found that such perturbations can be adapted into different model architectures, demonstrating that many deep neural networks are vulnerable to these input manipulations. Adversarial examples raise concerns about the trust one can place in neural network classifiers, and much work has been put into adversarial training methods to improve the robustness of models to adversarial examples. The most popular adversarial training regimes [1] generate adversarial examples (with corrected labels) within a fixed "attack radius" ($\epsilon$) during training procedure and include them in the model's training dataset. While adversarial training has been shown to increase robustness to adversarial examples drastically, it often comes at some cost to standard accuracy [28].

There is an inherent contention between the considerations of algorithmic recourse and adversarial robustness. While minimizing the changes necessary to alter a classifier's decision is seen as beneficial from a recourse perspective, such changes are harmful from a robustness perspective. Research [14] has demonstrated that adversarial training increases the average recourse cost, with higher adversarial training radii corresponding to higher recourse costs, which raises the concern that there may be an inherent trade-off between robustness and recourse.

Briefly, it should be noted that the goals of adversarial robustness are not totally at odds with recourse. Recourse *should* represent *true movements towards a desired class*, and adversarial examples that "fool" a model can be *harmful* and should *not be presented as recourse*. Consider the lending setting: if an approval action plan is provided to an applicant which does not represent a true movement in their underlying propensity for repayment, both the lender and borrower are putting themselves at long-term financial risk by following that plan. This is relevant in the context of many recourse settings, where data is tabular and it is not immediately obvious which input perturbations constitute adversarial examples and which input perturbations constitute recourse that genuinely moves an individual towards a desired class manifold. With this in mind, it is worth considering not only the change in overall cost of recourse, but also the change in proximity of recourse to the desired data manifold, when selecting an adversarial training radius.

---

*Corresponding to yangliu@ucsc.edu.

Even more fundamentally, it is important to question *whether a fixed adversarial training radius is appropriate, particularly in the context of algorithmic recourse?*. It has been shown [2] that different data instances have different *inherent adversarial vulnerabilities* due to their varying proximities to other classes. As such, some researchers have argued that an identical adversarial training radius should not be applied to all instances during training. Several methods [2, 6, 8] have been proposed for automatically learning *instance-wise* adversarial radii to address this variability. These are broadly referred to as "Adaptive Adversarial Training" (AAT) regimes [1].

This work explores the effects of AAT on both model robustness and ultimate recourse costs in an attempt to address the trade-off between the two and find a *justifiable* middle ground. Our contributions include:

- An observation on the effects of robustness on recourse costs, and when AAT yields more affordable recourse.
- Experiments demonstrating AAT's superior robustness/ recourse trade-offs over traditional AT.

## 2 BACKGROUND AND RELATED WORKS

*Algorithmic Recourse:* The continued adoption of ML in high-impact decision making such as banking, healthcare, and resource allocation has inspired much work in the field of Algorithmic Recourse [11, 13, 24], and Counterfactual Explanations [15, 19, 21, 27]. The performance of different recourse methods depends highly on properties of the datasets they are applied to, the model they operate on, the application of that model's score, and factual point specificities [7]. However, broadly speaking, recourse methods are classified based on: i) the *model family* they apply to, ii) the degree of *access* they have to the underlying model (i.e. white vs. black box methods), iii) the consideration of *manifold proximity* in the generation of recourse, iv) the underlying *causal relationships* in the data, and v) the use of *model approximations* in the generation process [26]. Recently, [18] introduced CARLA, a framework for benchmarking different recourse methods which act as an aggregator for popular recourse methods and standard datasets.

*Adversarial Attacks and Adversarial Training:* Adversarial vulnerability refers to the susceptibility of a model to be *fooled* by perturbations to the input data which cannot be detected by humans (so-called *Adversarial Examples*) [23]. Adversarial Training [10, 16] has been introduced to create models which are not susceptible to such attacks. The most popular method of Adversarial Training generates adversarial examples during the training process and includes them in the training dataset with corrected labels alongside the uncorrupted dataset. Often, adversarial training comes at some cost to standard classification accuracy. There have been many attack methods proposed to generate adversarial examples [5] with varying degrees of access to the model under attack, but most focus on defending against adversarial examples within a given $\epsilon$-radius (which are often defined by $\ell_1$, $\ell_2$, or $\ell_\infty$ norms of size $\epsilon$.) This work follows the popular attack and training formulation from [16], which minimizes the worst-case loss within a defined $\epsilon$-radius.

*On the Intersection of Robustness and Recourse.* Both Adversarial Examples and Counterfactual Explanations are formally described as constrained optimization problems where the objective is to alter a model's output by minimally perturbing input features [4, 9]. Recent work [17] proved equivalence between certain adversarial attack methods and counterfactual explanation methods, and further work has demonstrated both theoretically and empirically that increasing the radius of attack during adversarial training increases the cost of the resulting recourse [14]. This inherent connection pits security at odds with expressivity and raises an important question as to how an adversarial radius ought to be selected for adversarial training. If the radius is too small, the model may be overly sensitive to an attack, while if it is too large, end users suffer from potentially overly-burdensome recourse costs. In the context of many recourse problems where data is tabular, it is difficult to determine what may constitute an adversarial attack, furthering the difficulty of radius selection. [3] discussed a formulation for adversarial attacks on tabular data that accounts for both the radius of attack and the importance of a feature, but this is difficult to know a priori and often changes depending on the choice of explanation method selected [20].

*Adaptive Adversarial Training.* It has been observed that different data instances have different inherent adversarial vulnerability due to their varying proximity to other class' data manifolds, calling into question the conventional wisdom that models should be adversarially trained at a single consistent adversarial radius. [2] first observed this issue in the image classification domain, where certain instances can be *meaningfully transformed* into other classes even at small adversarial radii. The authors of [2] proposed a means of discovering instance-wise adversarial radii by iteratively increasing or decreasing each instance's attack radius based on whether attacks are successful. [6] built on this work by further motivating the effects of overly-large adversarial radii on classification accuracy and proposed a variation of [2]'s method which included adaptive label-smoothing to account for the uncertainty added by larger attack radii, and [8] proposed a means for adaptive adversarial training by increasing the classification margin around correctly-classified datapoints. Adaptive Adversarial Training (AAT) presents a means of "automatically" selecting attack radii during training, and in all works thus far, has shown positive results in terms of the accuracy/robustness trade-off inherent in adversarial training, as well as smoother robustness curves across ranges of attack radii compared with traditional Adversarial Training.

## 3 PRELIMINARIES & NOTATION

*Standard Training:* We begin with a model $f$ parameterized by weights $\theta$ that maps $X \to Y$, where $x \in X$ are features and $y \in Y$ are their corresponding labels. Given a dataset $\mathcal{D} = \{(x_i, y_i)\}_{i=1}^{N}$, and a loss function $\ell(\cdot)$, a standard learning objective is to minimize the average loss on the data:

$$\min_{\theta} \quad \frac{1}{N} \sum_{(x_i, y_i) \in D} \ell(f_\theta(x_i), y_i) \tag{1}$$

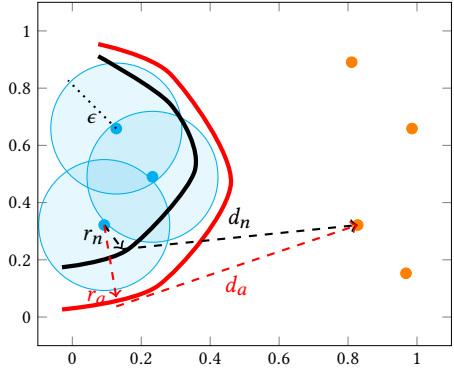

(a) Toy problem demonstrating that adversarial training can result in counterfactuals that are both costlier and further from the desired class manifold. The natural decision boundary is shown in black, the adversarial boundary in red. $\epsilon$-Adversarial training creates a necessary recourse cost $c_a = \epsilon > c_n$, and yields a distance in the resulting recourse to the desired manifold of $d_a > d_n$

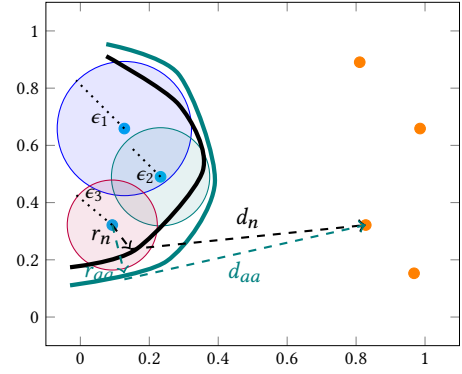

(b) Adaptive Adversarial Training provides counterfactuals which are cheaper and relatively closer to the desired class manifold. The natural decision boundary is shown in black, the adaptive adversarial boundary in green. With instance specific robustness $\epsilon_i$, the recourse cost $c_{aa} = \epsilon_i > c_n$ and $c_{aa} < \epsilon$ for any $\epsilon_i < \epsilon$. This yields a distance $d_{aa} < d_a$.

Figure 1: An example scenario demonstrating the effectiveness of AAT in terms of recourse costs.

Let $f_{nat}$ represent the naturally trained model using the standard loss minimization based optimization technique.

*Adversarial Attacks:* The goal of an adversarial attack is to strategically generate perturbations $\delta$ which can significantly enlarge the loss $\ell(\cdot)$ when added to an instance $x$. [10] introduced *Fast Gradient Sign Method (FGSM)* for generating adversarial examples using the following mechanism:

$$x'_i = x_i + \alpha \cdot \text{sign}\left(\nabla_{x_i}\ell\left(f_\theta(x_i), y_i\right)\right) \qquad (2)$$

where $\alpha$ denotes the size of the perturbation, $x'_i$ denotes the adversarially perturbed sample, and $x_i$ is the original clean sample. The *sign* function operates on the gradient of $\ell\left(f_\theta(x_i), y_i\right)$ w.r.t. $x_i$, which is used to set the gradient to 1 if it is greater than 0 and −1 if it is less than 0. [16] proposed a stronger iterative version of FGSM, performing Projected Gradient Descent (PGD) on the negative loss function:

$$x_i(t+1) = \Pi_{x+S}\left(x_i(t) + \alpha \cdot \text{sign}\left(\nabla_{x_i(t)}\ell(f_\theta(x_i(t)), y_i)\right)\right)$$

where $\alpha$ denotes the perturbation step size at each iteration and $x_i(t+1)$ represents the perturbed example at step $t+1$ for the clean instance $x_i$. In this work, we use PGD due to its performance, popularity, and relative speed.

*Adversarial Training:* Adversarial training is usually formulated as a min-max learning objective, wherein we seek to minimize the worst case loss within a fixed training radius $\epsilon$.

$$\min_\theta \max_{||\delta_i|| \leq \epsilon} \frac{1}{N} \sum_{(x_i, y_i) \in D} \ell(f_\theta(x_i + \delta_i), y_i) \qquad (3)$$

We solve this min-max objective via an alternating stochastic method that takes minimization steps for $\theta$, followed by maximization steps that approximately solve the inner optimization using $k$ steps of an adversarial attack. PGD with a fixed $\epsilon$ is used to perturb an original instance and let $f_{\epsilon\text{-adv}}$ represent the model trained with a PGD radius of $\epsilon$.

### 3.1 Adaptive Adversarial Training

[2] first argued that different data instances have different intrinsic adversarial vulnerabilities due to their varying proximity to other class manifolds, and introduced Instance-Adaptive Adversarial Training (AAT) to automatically learn instance-wise adversarial radii. The authors proposed the following objective function:

$$\min_\theta \max_{||\delta_i|| \leq \epsilon_i} \frac{1}{N} \sum_{(x_i, y_i) \in D} \ell(f_\theta(x_i + \delta_i), y_i) \qquad (4)$$

where $\epsilon_i$ denotes each training instance's attack radius. $\epsilon_i$ is iteratively updated at each training epoch, increasing by a constant factor if the attack at the existing radius is unsuccessful and decreasing by a constant factor if it is successful.

[8] presented an alternate form of AAT called Max-Margin Adversarial (MMA) Training that seeks to impart adversarial robustness by maximizing the margin between correctly classified datapoints and the model's decision boundary. Formally, they proposed the following objective:

$$\min_\theta \left\{ \sum_{i \in S_\theta^+} \max\{0, d_{max} - d_\theta(x_i, y_i)\} + \beta \sum_{i \in S_\theta^-} \ell(f_\theta(x_j), y_j) \right\}$$
(5)

where $S_\theta^+$ is the set of correctly classified examples, $S_\theta^-$ is the set of incorrectly classified examples, $d_\theta(x_i, y_i)$ is the margin between correctly classified examples and the model's decision boundary, $d_{max}$ is a hyper-parameter controlling which points to maximize the boundary around (forcing the learning to focus on points with $d_\theta$ less than $d_{max}$,) and $\beta$ is a term controlling the trade-off between standard loss and *margin maximization*. The authors use a line search based on PGD to efficiently approximate $d_\theta(x_i, y_i)$. For the rest of this study, let $f_{aat}$ be a model trained using a mechanism from this category of training techniques.

## 3.2 Recourse Methods

For the scope of this study, we explore three different classes [14] of recourse methods: i) one random search, ii) one gradient-based search, and iii) one manifold-based approach. We will now briefly discuss each method, and we refer the readers to the original works for further implementation details.

*Growing Spheres (GS):.* [15] proposed a random search method for calculating counterfactual by sampling from points within $\ell_2$-hyper-spheres around $x$ of iteratively increasing radii until one or more counterfactual is identified which flips $f(x)$. Formally, they present a minimization problem in selecting which counterfactual $x'$ to return:

$$\arg\min_{x' \in \mathcal{X}} \{c(x, x') | f(x) \neq f(x')\} \quad (6)$$

where $\mathcal{X}$ is the family of sampled points around $x$ and $c$ is a cost function in $\mathcal{X} \times \mathcal{X} \to \mathbb{R}_+$: $||x' - x||_2 + \gamma ||x' - x||_0$, where $\gamma$ is a hyperparameter controlling the desired sparsity of the resulting counterfactual.

*Score Counterfactual Explanations (SCFE):.* [27] proposed a gradient-based method for identifying counterfactuals $x'$.

$$\arg\min_{x'} \max_{\lambda} \lambda(f(x') - y')^2 + d(x, x') \quad (7)$$

where $d(\cdot, \cdot)$ is some distance function and $y'$ is the desired score from the model. In practice, this is solved by iteratively finding $x'$ and increasing $\lambda$ until a satisfactory solution is identified.

*CCHVAE:.* [19] proposed a manifold-based solution to finding counterfactuals using a Variational Auto Encoder (VAE) to search for counterfactuals in a latent representation $\mathcal{Z}$. The goal of CCHVAE and other manifold methods is to find counterfactuals that are semantically "similar" to other data points. Formally, given an encoder $\mathcal{E}$, a decoder $\mathcal{H}$, and a latent representation $\mathcal{Z}$ where $\mathcal{E}: \mathcal{X} \to \mathcal{Z}$, CCHVAE optimizes the following:

$$\arg\min_{z' \in \mathcal{Z}} \{||z'|| \; s.t. \; f(\mathcal{H}(\mathcal{E}(x) + z')) \neq f(x)\} \quad (8)$$

## 4 RECOURSE TRADE-OFFS WITH ADAPTIVE ADVERSARIAL TRAINING

*Recourse cost.* The cost of recourse is usually approximated using a distance based metric. A common practice among recourse methodologies is to minimize the cost in some form or the other, because in general a low cost recourse is assumed to be easier to act upon. The cost of a recourse for a classification based model is traditionally interpreted as the minimum distance between a factual and the decision boundary. Alternatively, the inherent goal of adversarial training is to maximize the distance between factuals and the decision boundary. Hence, traditional adversarial training exacerbates the recourse costs of a classifier. In this section, we make preliminary observations on the effects of adaptive adversarial training on recourse costs.

An increase in $\epsilon$ for $\epsilon$-adversarial training increases the overall recourse costs and the corresponding relation between $\epsilon$ and $C$ is discussed in [14]. In comparison with an $\epsilon$-adversarial training, we observe the following benefits from the instance adaptive adversarially training:

## 4.1 Recourse Costs

Let $\delta_x^{(nat)} = d(x, x')$ be the distance to the closest adversarial example $x'$ for the instance $x$ for a standard training based model, and, analogously, let $c_x^{(nat)} = cost(x, x'')$ be the cost of a recourse $x''$ for an individual represented by $x$. For simplicity, we assume that both $c_{(\cdot)}^{(\cdot)}$ and $cost(\cdot, \cdot)$ use the same $\ell_p$ norm based distance metrics. Let $H^- = \{x \in \mathcal{X} : f(x) = -1\}$ represent the sub-population which was adversely affected by the classifier $f(\cdot)$, and analogously we have $H^+ = \{x \in \mathcal{X} : f(x) = +1\}$. The average cost of recourses for $H^-$ is defined for a naturally trained model as:

$$c_*^{(nat)} = \frac{1}{|H^-|} \sum_{x \in H^-} c_x^{(nat)} \quad (9)$$

Let $\underline{H}^- = \{x \in \mathcal{X} : f(x) = -1, c_x^{(nat)} \leq \underline{\epsilon}\}$, where $\underline{\epsilon}$ is a cost threshold to identify low cost recourses. As observed in Figures 4 and 5, a low cost counterfactual is sufficient in practice for a large section of the population. However, an optimal $\epsilon_a$-adv classifier provides at least $\epsilon_a$ robustness to all samples in the training dataset. This can be visualized by the sharp peak in the distribution of the observed $\epsilon$ in the test dataset for all the $\epsilon$-adv models (Figure 8). However AAT models provide natural robustness to the data samples, meaning that a data instance closer to the natural decision boundary has $\epsilon_{aat}^{H^-}$ that depends on the data's natural proximity to the decision boundary. For instances with $\epsilon_{aat}^{H^-} < \epsilon_a$, the resulting recourse will be more affordable. For $\epsilon_{aat}^{H^-} < c_x^{(nat)}$, low cost recourse within $\underline{H}^-$ will be preserved.

## 4.2 Proximity to the Desired Manifold

*Manifold Proximity* measures the distance by some metric between recourse and the target sub-population. For an $f_{\epsilon_a\text{-adv}}^*$ model, the recourse suggested have at least $\epsilon_a$ proximity from the target approved sub-population $H^+$ due to the fact that the target sub-population is also $\epsilon_a$ away from the decision boundary. Alternatively $f_{aat}$ is naturally robust for the target sub-population as well. Hence, the Recourse provided has the potential to be closer in terms of proximity to $H^+$, so long as $\epsilon_{aat}^{H^+} < \epsilon_a$. We report the average proximity $\rho_{f_{\epsilon\text{-adv}}}$ of the model $f_{\epsilon\text{-adv}}$ using:

$$\rho_{f_{\epsilon\text{-adv}}} = \frac{1}{|N_{test}|} \sum_{x \in N_{test}} \min_{x^+ \in H^+} d(x, x^+) \quad (10)$$

where $d(x, x^+)$ is a distance measure between a counterfactual $x$ and a target population $x^+$. We report both $\rho_{f_{\epsilon\text{-adv}}}$ and $\rho_{f_{aat}}$ for the corresponding models. In Figure 7, we find that $\rho_{f_{aat}}$ is significantly better than $\rho_{f_{\epsilon\text{-adv}}}$. A motivating toy problem demonstrating lower recourse costs and closer manifold proximity is also visualized in Figure 1.

## 4.3 Preservation of Low Cost Recourse

The recourse costs provided to the adversely affected individuals by a model should follow the natural distribution of the difficulty of acting upon the suggested recourse at the population level. With a fixed $\epsilon$ while training an optimal adversarially trained $f_{\epsilon\text{-adv}}^*$ model, the recourse suggested must necessarily be $\epsilon$ away from the decision boundary and further $\epsilon$ away from the nearest target population sample. Such counterfactuals contradicts with the recourse literature [25], which describes a distribution in recourse costs wherein a proportion of individuals only require minimal low cost actionable steps to obtain the desired outcome from a model, whereas other individuals can have a much larger recourse costs. Essentially, $\epsilon$-robustness necessarily denies recourse with lower costs than $\epsilon$.

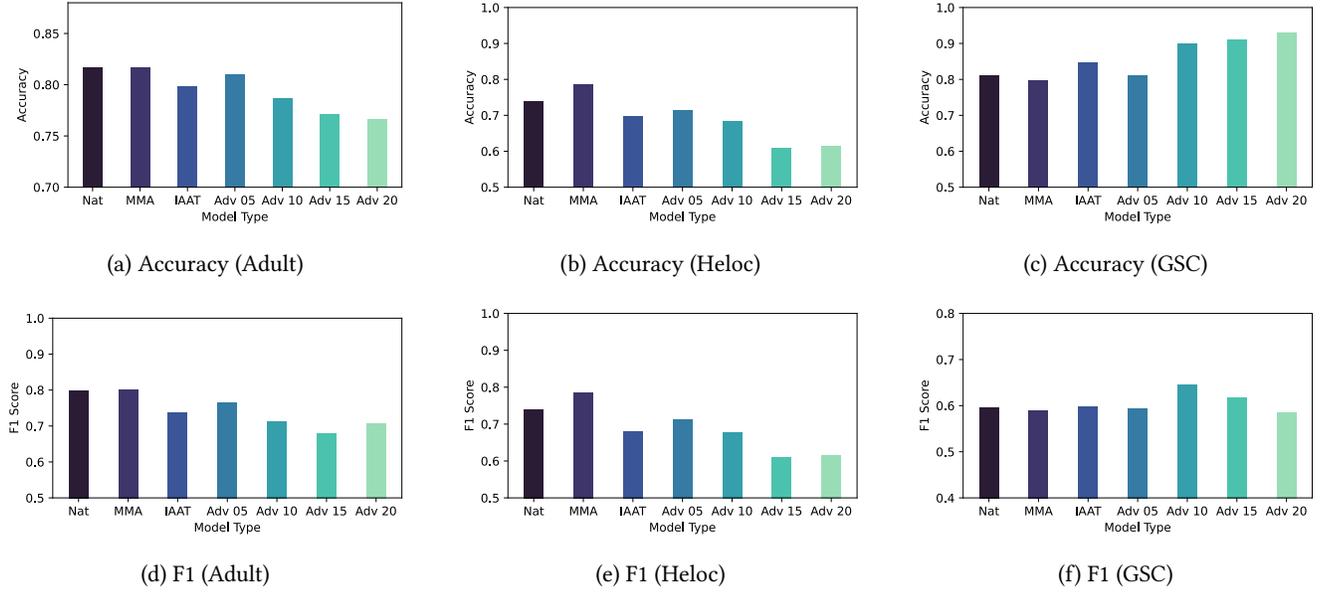

Figure 2: Standard performance across datatsets. MMA shows particularly competitive standard performance compared with all other Adversarial Training regimens.

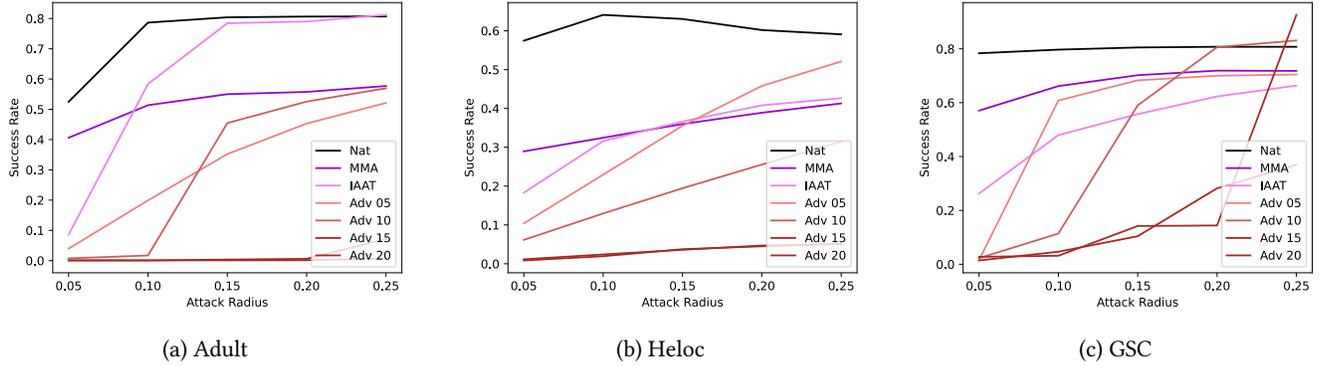

Figure 3: Attack Success Rate. Traditional Adversarial Training shows higher robustness within its predefined training threshold, but sharper robustness degradation as the attack radius increases.

$f_{aat}$ does not enforce a strict $\epsilon$ while training, allowing instances to have a wider range of recourse costs. To this end we compare the rate of extreme low cost recourse $C_\Delta$ across the discussed training methods with real-world datasets to measure the rate at which it degrades in practice. For simplicity, we measure:

$$C_\Delta = \frac{1}{|N_{test}|} \sum_{x_i \in N_{test}} \mathbf{1}(C_{x_i} < \epsilon) \qquad (11)$$

where $C_{x_i}$ is the cost of recourse for an instance $x_i$ and $\epsilon$ is a minimum adversarial training radius. We observe in Figure 4 that Adaptive Adversarial Training preserves low cost recourse rates despite providing overall robustness benefits.

## 5 EXPERIMENTAL DESIGN & METRICS

In this section, we detail our experimentation procedure to empirically evaluate these various training methods and explain our metric choices. The CARLA package [18] was used to source the datasets and recourse methods we employed.

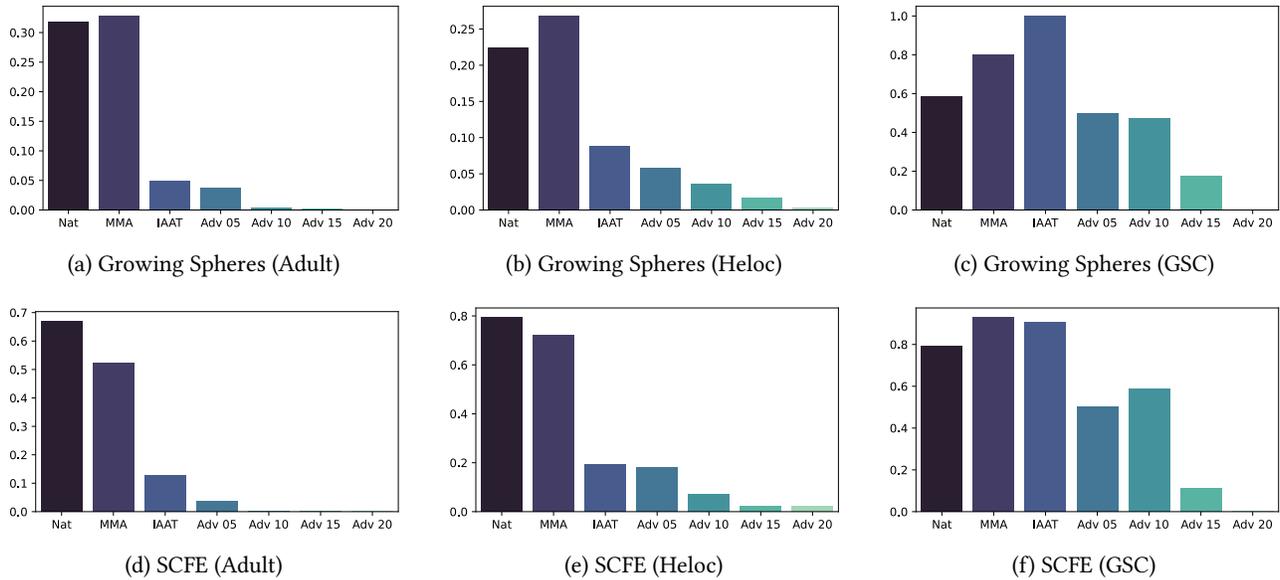

Figure 4: Low cost recourse ($\ell_\infty < 0.05$) proportion for methods that optimize directly in the input space. We observe that AAT models has much higher proportions of low cost recourse, supporting the hypothesis that it allows for robustness while preserving low recourse costs for individuals near natural decision boundaries.

## 5.1 Experimental Setup

*Datasets.* We performed our experiments on three datasets:

- *Adult Income*: A dataset originating from the 1994 Census of 48,842 individuals for whom the task is to predict whether someone makes more than $50,000/yr. It is comprised of 20 features which are a combination of demographic features (age, sex, racial group), as well as employment features (hours of work per week and salary), and financial features (capital gains/losses.) In keeping with [14] and [3], we removed categorical features for efficient training and approximation of tabular adversarial examples. The target distribution is somewhat skewed, with a 76% positive label proportion.
- *Home Equity Line of Credit (Heloc)*: pulled from the 2019 FICO Explainable Machine Learning (xML) challenge, the Heloc dataset consists of anonymized credit bureau data from 9,871 individuals where the task is to predict whether an individual will repay their HELOC account within two years. The dataset consists of 21 financial features and no demographic data. The target distribution is evenly split, with a 48% positive label proportion.
- *Give Me Some Credit (GSC)*: a credit-scoring dataset pulled from a 2011 Kaggle Competition consisting of 150,000 individuals for whom the task is to predict default. It consists of 11 features, one of which is a demographic feature (age), and the rest are financial variables. The target distribution is heavily skewed, with a 93% positive label proportion.

*Models.* We trained a total of 7 Neural Network models for each of our datasets: one naturally trained model, one model trained with AAT, one model trained with MMA, and four adversarially trained models. All models are trained using Binary Cross Entropy with the default model architecture from CARLA, with three hidden layers of [18, 9, 3] units. The Adversarially Trained models were all trained with PGD at a variety of $\epsilon \in [0.05, 0.1, 0.15, 0.2]$. The AAT model did not consider any hyperparameter choices, and the MMA model was trained using the original work's package [8] with the default hyperparameter choices.

*Recourse Methods.* We constructed Counterfactual Explanations for all models on a sample of 1000 negatively-classified test data points using three methods: Growing Spheres (GS), C-CHVAE, and SCFE. All hyperparameter choices for these methods were left as their CARLA defaults.

## 5.2 Metrics

To study the effects of the different training methods on accuracy, robustness, and recourse, we calculate the following metrics:

*Standard Classification Performance.* A primary consideration in adversarial training is the trade-off in classification accuracy when compared with natural training. We record the standard classification accuracy of all models to measure the drop in accuracy that may accompany the different adversarial training methods. Formally, we measure: $\frac{1}{|\mathcal{D}_{test}|} \sum_{x_i \in \mathcal{D}_{test}} \mathbf{1}(f(x_i) = y_i)$. Given that we are experimenting with datasets with skewed target distributions, we also record the F1 score of each model on the minority target population.

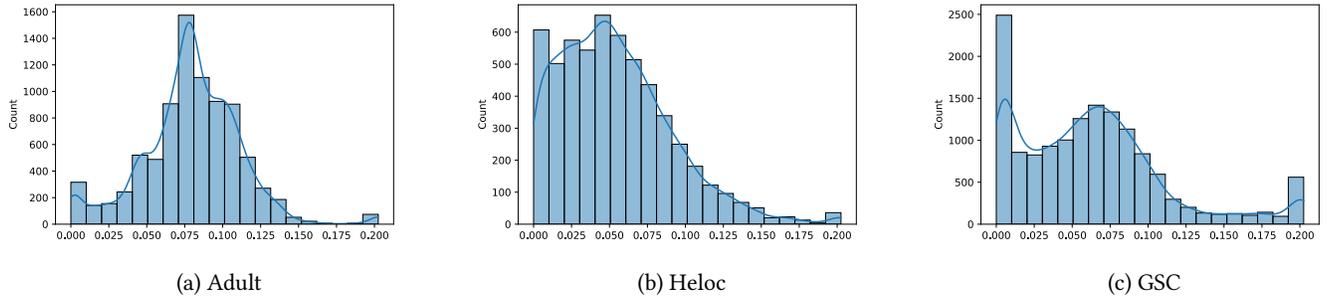

(a) Adult      (b) Heloc      (c) GSC

Figure 5: AAT "Discovered" Radii Resulting from Adpative Adversarial Training

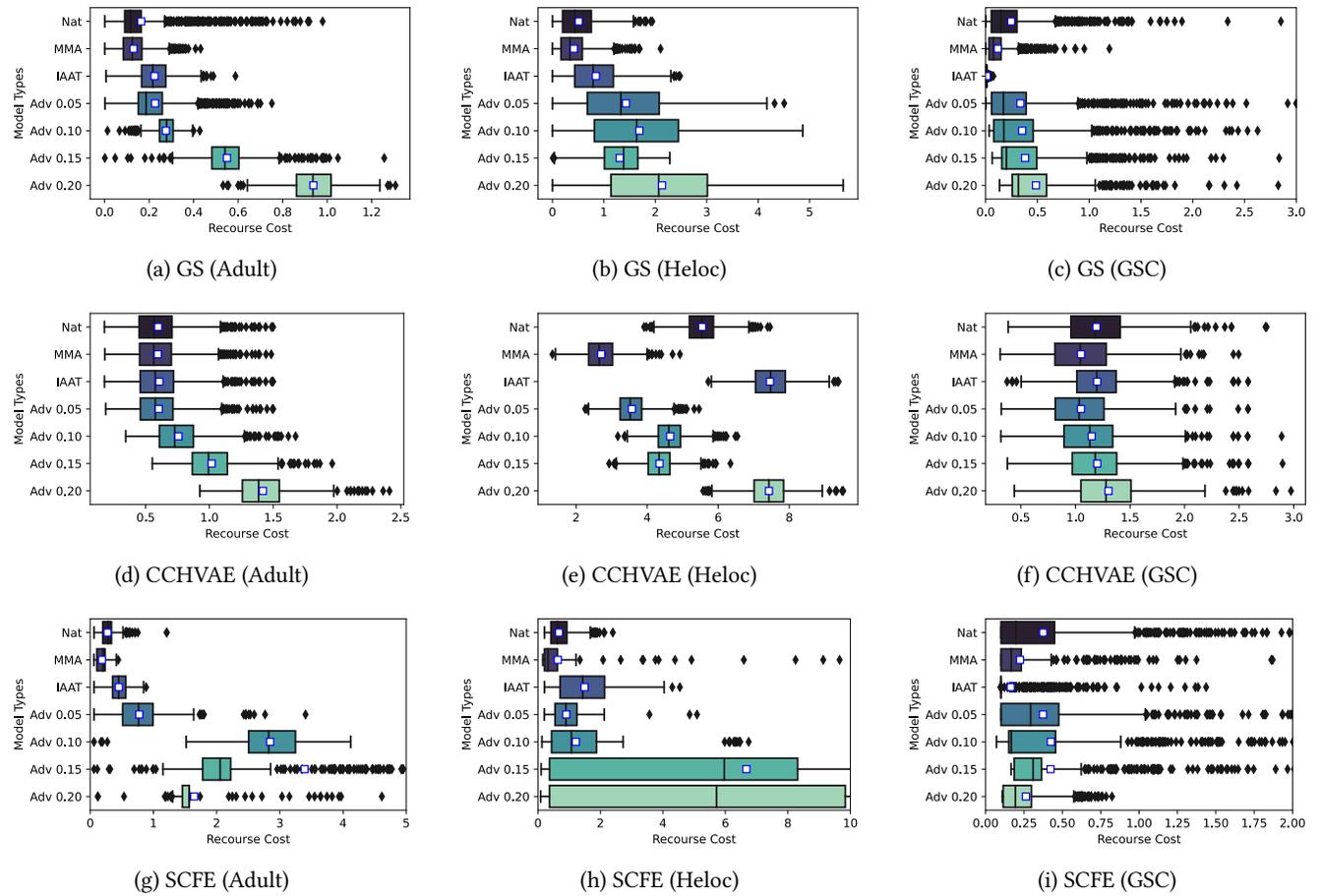

(a) GS (Adult)      (b) GS (Heloc)      (c) GS (GSC)

(d) CCHVAE (Adult)   (e) CCHVAE (Heloc)   (f) CCHVAE (GSC)

(g) SCFE (Adult)     (h) SCFE (Heloc)     (i) SCFE (GSC)

Figure 6: Recourse costs (defined as the $\ell_2$ distance between a factual and counterfactual data point) for all methods and datsets. We observe that adaptive adversarial training shows significantly more competitive recourse costs than traditional adversarial training, and MMA training in particular shows almost no increase over natural training despite its robustness benefits.

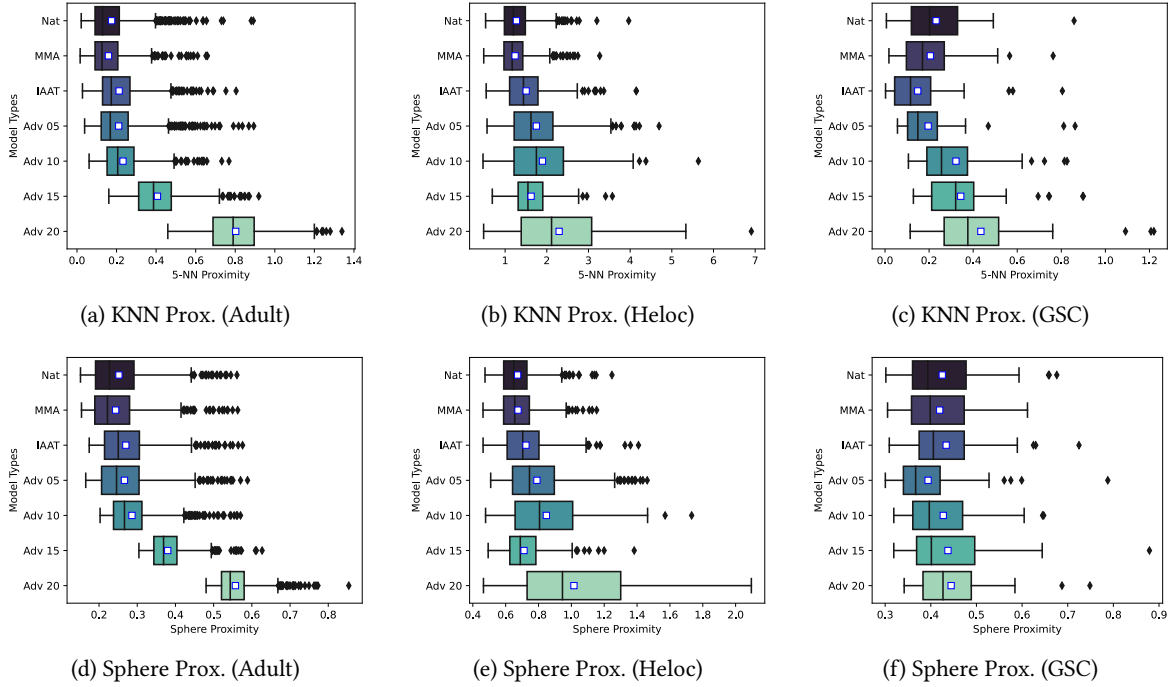

Figure 7: KNN and Sphere Manifold Proximity for Growing Spheres. We find that not only does adaptive adversarial training produce less expensive recourse than traditional adversarial training, but also recourse that is more faithful to the desired class these counterfactuals approximate.

*Adversarial Success Rate.* Given that we are primarily concerned with the trade-off between robustness and recourse, and following the concept of "boundary error" introduced in [29] to disentangle standard performance and adversarial vulnerability, we also measure the success rate of adversarial attacks at various radii on our models. Formally, given an attack $\mathcal{A}_\epsilon$ such that $\mathcal{A}_\epsilon(x)$ identifies the most adversarial example on $x$ within a radius $\epsilon$, we measure $\frac{1}{|\mathcal{D}_{test}|} \sum_{x_i \in \mathcal{D}_{test}} \mathbf{1}(f(\mathcal{A}_\epsilon(x)) \neq f(x_i))$. We observe the adversarial success rate across the radii on which we train our traditional adversarial models. Note that this is an imperfect metric for measuring the success of AAT, as AAT assumes that some "attacks" at given radii represent real movements toward different classes; however, it is still useful to capture this information in considering the trade-off between traditional adversarial training and AAT.

*Counterfactual Proximity.* The primary metric regarding recourse we are interested in observing is the ultimate recourse cost between our resultant models. As each specific domain's cost function is not concretely defined, we follow the convention of opting for $\ell_2$ distance as a standard approximation. Formally, for each model we calculate: $\frac{1}{|\mathcal{D}_{test}|} \sum_{x_i \in \mathcal{D}_{test}} ||x_i^* - x_i||_2$, where $x^*$ is the recourse calculated for $x_i$.

*Manifold Proximity.* Motivated by the question of how faithful our resulting counterfactuals are to true movements towards the desired class, we estimate the distance between the counterfactuals each model produces and the desired class manifold these counterfactuals approximate. We use two methods for this: a KNN distance measure and a sphere distance measure For KNN, we record the average $\ell_2$ distance between the resulting counterfactuals and the five nearest neighbors of the desired class. For the sphere measure, we record the average $\ell_2$ distance between the resulting counterfactuals and all neighbors of the desired class within an $\ell_2$ ball of size $\epsilon$, where $\epsilon$ is calculated as 20% of the average $\ell_2$ distance between any two points in the dataset.

## 6 RESULTS & DISCUSSION

*Standard Performance.* Figure 2 displays the classification accuracy and F1 scores of the various models. We observe that for the Adult and Heloc datasets, adversarial training tends to decrease standard performance, with higher training radii correlating with worse performance. We observe that MMA training tends to keep performance consistent, and that AAT worsens performance to a degree similar to adversarial training with an $\epsilon$ value between 0.05 and 0.1.

*Robustness.* Figure 3 shows the vulnerability of the models under PGD attack at a variety of raddii ($\epsilon \in [0.05, 0.1, 0.15, 0.2, 0.25]$). We observe that while traditional adversarial training creates substantially more robust models within a defined radius of attack, the degradation in robustness tends to be more severe among traditionally trained models than AAT methods when the radius increases beyond their predefined training threshold. MMA in particular shows surprisingly consistent robustness benfits, although they are more moderate than their adversarially trained counterparts'.

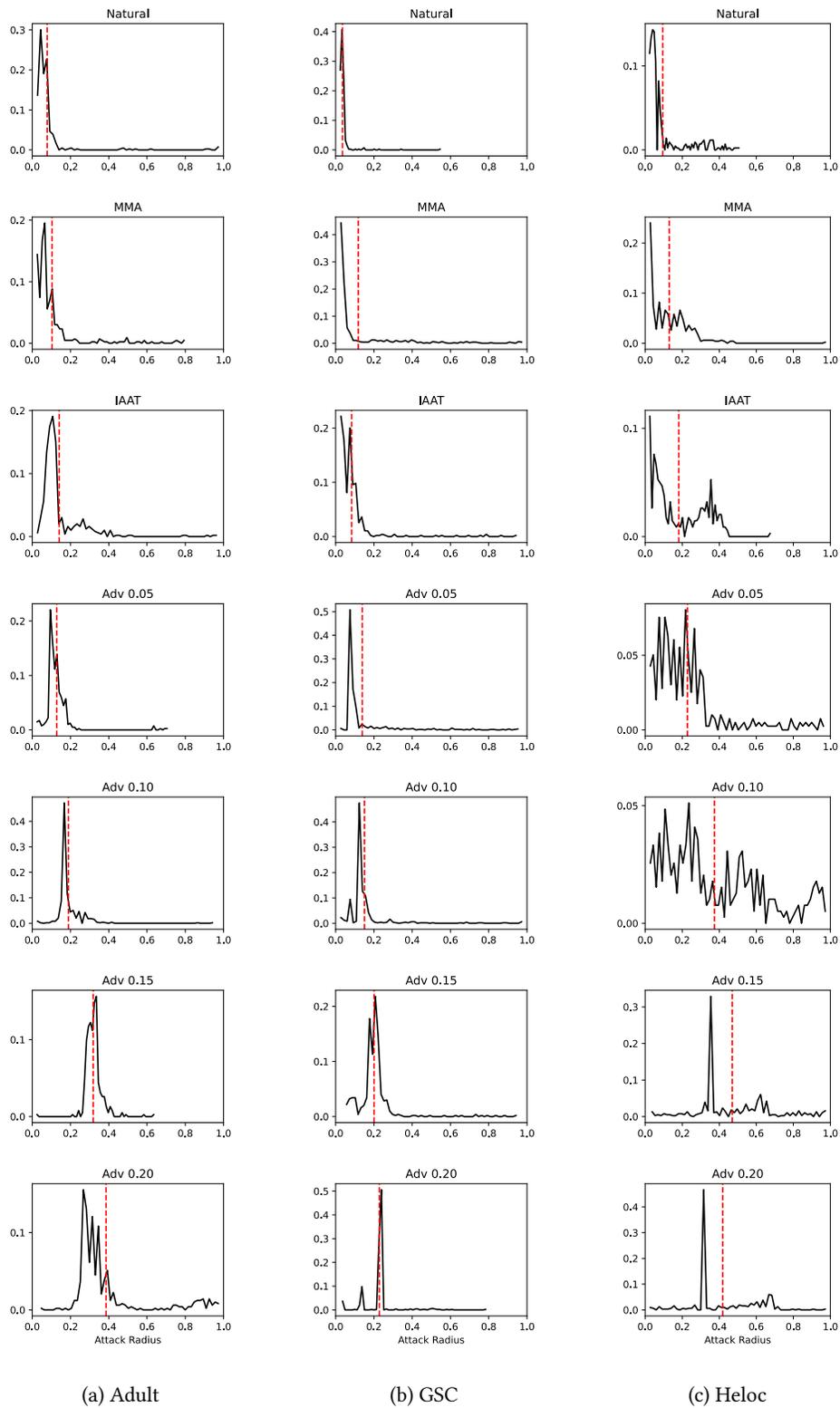

Figure 8: Decision boundary proximity, estimated by the minimum successful PGD attack radius on a sample of 1000 instances. The height represents a proportion of the data, the average distance is shown in red.

*Counterfactual Proximity.* Figure 6 displays the cost of recourse across all datasets for the three recourse methods studied. We observe consistently that adaptive adversarial training yields recourse with lower costs than traditional adversarial training, and in the case of MMA costs that are consistently competitive with natural training. This result seems unintuitive given the robustness benefits that MMA provides, and we believe this presents an interesting avenue for further research.

*KNN & Sphere Manifold Distance.* Figure 7 shows the Manifold Proximity estimates for Growing Spheres across all datasets. We observe that adaptive adversarial training produces recourse that is consistently closer to the desired class manifold than traditional adversarial training. This result, paired with the reduction in recourse costs, may suggest that adaptive adversarial training encourages more natural decision boundaries than traditional adversarial training, allowing for more meaningful recourse at lower costs.

*Prevalence of Low Cost Recourse.* For recourse methods that optimize costs directly in the input space, we record the percentage of counterfactuals that have an $\ell_\infty$ cost less than 0.05 to measure the proportion of low cost recourse among our models. The results are recorded in Figure 4. We observe that adaptive adversarial training shows higher proportions of low cost recourse than traditional adversarially trained models; surprisingly, MMA training in particular finds proportions of low-cost recourse that are consistently competitive with natural training, despite its benefits in overall robustness.

*Discovered Radii & Decision Boundary Distances.* Figure 5 displays the instance-wise discovered radii after AAT for all three datasets. We observe that for all datasets, a variety of radii are found with unique distributions. This alludes to the fact that different underlying data distributions have different levels of inherent adversarial vulnerability, underscoring the challenge of estimating a proper singular radius at which to adversarially train. Figure 8 shows an estimation of the distribution of decision boundary proximities across all models, calculated by finding the minimum successful radius for PDG attack across a sample of 1000 instances. We observe that traditional $\epsilon$-adversarial training often limits proximity to the decision boundary $d > \epsilon_i$, while adaptive adversarial training shows a greater distribution in ultimate decision boundary proximties. In the case of MMA in particular, we find that the decision boundary proximities closely match that of the natural model, despite its improved robustness.

## 7 CONCLUSION

This work explores the effects of adaptive adversarial training on robustness and recourse, finding that it shows promising trade-offs between the two. We motivate our work with a observation of the effect of traditional adversarial training on recourse costs, and introduce scenarios under which adaptive adversarial training provides more affordable recourse. We conduct experiments on three datasets demonstrating that adaptive adversarial training yields significant robustness benefits over natural training with little cost incurred on recourse and standard performance, and provide evidence that adaptive adversarial training produces recourse that more faithfully represents movements towards the desired class manifold. Finally we analyze the resulting models' decision boundary margins, providing evidence that supports our observations on recourse costs under traditional adversarial training. We believe that adaptive adversarial training, and Max-Margin adversarial training in particular, presents a promising means of achieving the ultimate goals of robustness while preserving affordable recourse costs for end users.

## ACKNOWLEDGMENTS

This work is partially supported by the National Science Foundation (NSF) under grants IIS-2143895 and IIS-2040800, and CCF-2023495.